\documentclass[runningheads]{llncs}
\usepackage[T1]{fontenc}
\usepackage{graphicx,verbatim}
\usepackage{svg}
\usepackage{caption}
\usepackage{booktabs}
\usepackage{multirow}
\usepackage{adjustbox}
\usepackage{arydshln}
\usepackage[colorlinks, citecolor=green]{hyperref}
\usepackage{amsmath}    
\usepackage{amssymb}    
\usepackage{amsfonts}   
\usepackage{array,graphicx,booktabs,multirow} 
\usepackage{caption,subcaption}
\usepackage{colortbl}
\usepackage[table]{xcolor}

\begin{document}
%
\title{TumorFlow: Physics-Guided Longitudinal MRI Synthesis of Glioblastoma Growth} 

\titlerunning{Physics-Guided Longitudinal MRI Synthesis of Glioblastoma Growth}
%
\author{Valentin Biller \inst{1}*\and
Niklas Bubeck \inst{2,4}* \and
Lucas Zimmer \inst{3} \and
Ayhan Can Erdur \inst{4} \and
Sandeep Nagar \inst{3} \and
Anke Meyer-Baese \inst{6} \and
Daniel Rückert \inst{2,4,5} \and
Benedikt Wiestler \inst{2,3}** \and
Jonas Weidner \inst{2,3}** 
}
\authorrunning{V. Biller et al.}
\institute{
Technical University of Munich (TUM), Germany \and
Munich Center for Machine Learning (MCML) \and
AI for Image-Guided Diagnosis and Therapy, TUM, Germany \and
Chair for AI in Healthcare and Medicine, TUM and TUM University Hospital, Munich, Germany \and
Imperial College London \and 
Florida State University \\
* equal contribution \\
** equal contribution \\
\email{\{valentin.biller, niklas.bubeck, j.weidner\}@tum.de} 
}


\maketitle              
\begin{abstract}

Glioblastoma exhibits diverse, infiltrative, and patient-specific growth patterns that are only partially visible on routine MRI, making it difficult to reliably assess true tumor extent and personalize treatment planning and follow-up.
We present a biophysically-conditioned generative framework that synthesizes biologically realistic 3D brain MRI volumes from estimated, spatially continuous tumor-concentration fields. Our approach combines a generative model with tumor-infiltration maps that can be propagated through time using a biophysical growth model, enabling fine-grained control over tumor shape and growth while preserving patient anatomy.
This enables us to synthesize consistent tumor growth trajectories directly in the space of real patients, providing interpretable, controllable estimation of tumor infiltration and progression beyond what is explicitly observed in imaging.
We evaluate the framework on longitudinal glioblastoma cases and demonstrate that it can generate temporally coherent sequences with realistic changes in tumor appearance and surrounding tissue response. These results suggest that integrating mechanistic tumor growth priors with modern generative modeling can provide a practical tool for patient-specific progression visualization and for generating controlled synthetic data to support downstream neuro-oncology workflows. In longitudinal extrapolation, we achieve a consistent 75\% Dice overlap with the biophysical model while maintaining a constant PSNR of 25 in the surrounding tissue. Our code is available at: \url{https://github.com/valentin-biller/lgm.git}

\keywords{Glioblastoma \and Tumor Growth Modeling \and Longitudinal Brain MRI Synthesis \and Biophysical Conditioning}

\end{abstract}

\begin{figure}[t]
        \centering
        \includegraphics[width=1.0\textwidth]{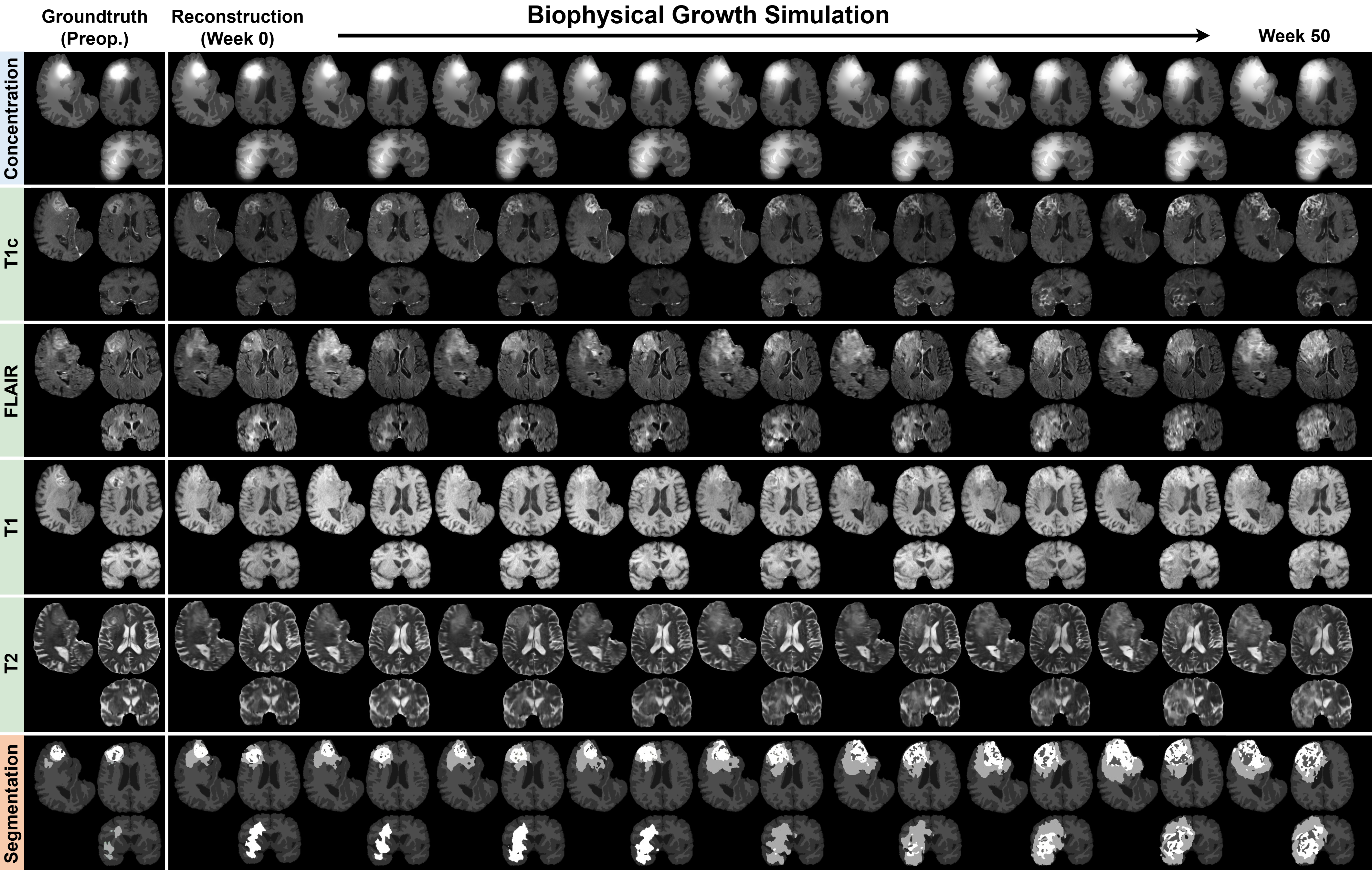}
        \captionof{figure}{\textbf{Representative longitudinal tumor growth prediction}. Starting from the preoperative state (week 0), a biophysical growth model is used to simulate time-resolved tumor concentration fields (blue), which serve as conditioning for image synthesis. The model generates T1c, FLAIR, T1, T2 modalities (green), preserving anatomy while capturing modality-specific progression. A standard segmentation model applied to the generated images yields time-consistent masks (orange).}
        \label{fig:synthetic_progression} 
\end{figure}

\section{Introduction}
Glioblastoma is characterized by a highly irregular, infiltrative growth pattern. Magnetic resonance imaging (MRI) is essential for diagnosis, treatment planning, and longitudinal monitoring. However, routine MRI fails to detect the true extent and spread of glioblastoma and is, in particular, blind to the diffuse infiltration of tumor cells into the surrounding brain \cite{yan2019multimodal}.

Biophysical brain tumor modeling has emerged as an interpretable and robust approach for estimating patient-specific tumor cell distributions and growth trajectories, enabling counterfactual predictions of tumor progression \cite{lipkova2019personalized,balcerak2025individualizing,weidner2024learnable}. These models, however, typically operate on a continuous tumor-concentration field, which is disconnected from the imaging space used in clinical practice for decision-making.

Recently, generative models have emerged for brain tumor MRI \cite{wang20253d,dorjsembe2024conditional,truong2024synthesizing,jiang2023cola}. However, they are conditioned on tumor segmentation masks, which constrain control to coarse morphology and prevent modeling spatially continuous infiltration patterns or their smooth evolution over time, a key requirement for \textit{in silico} modeling of tumor response and personalized treatment decisions.
In parallel, methods that are conditioned only on tumor volume were developed \cite{wolleb2022swiss,kebaili20243d,laslo2025mechanistic}. These approaches aim to implicitly infer biophysical growth characteristics from data, but they often fail to produce temporally coherent progression and lack modifiability (e.g., for simulating different treatments). This limitation is further exacerbated by the scarcity and heterogeneity of longitudinal datasets.

To enable realistic, interpretable estimation of longitudinal disease progression, we combine, for the first time, a biophysical tumor growth model with high-fidelity MRI generation.
Our main contributions are: \textbf{(i)} We train a generative flow matching model for 3D glioblastoma MRI on preoperative volumes only, which outperforms state-of-the-art approaches on image- and feature-level metrics.
\textbf{(ii)} We condition the model on a tumor infiltration field, enabling fine-grained control over tumor generation significantly beyond what classical segmentation masks offer.
\textbf{(iii)} We combine the generative model with state-of-the-art biophysical brain tumor modeling to generate robust and physically coherent longitudinal growth trajectories from cross-sectional MRI alone.

\section{Method}
\begin{figure}[t]
\begin{center}
\includegraphics[width=1.0\textwidth]{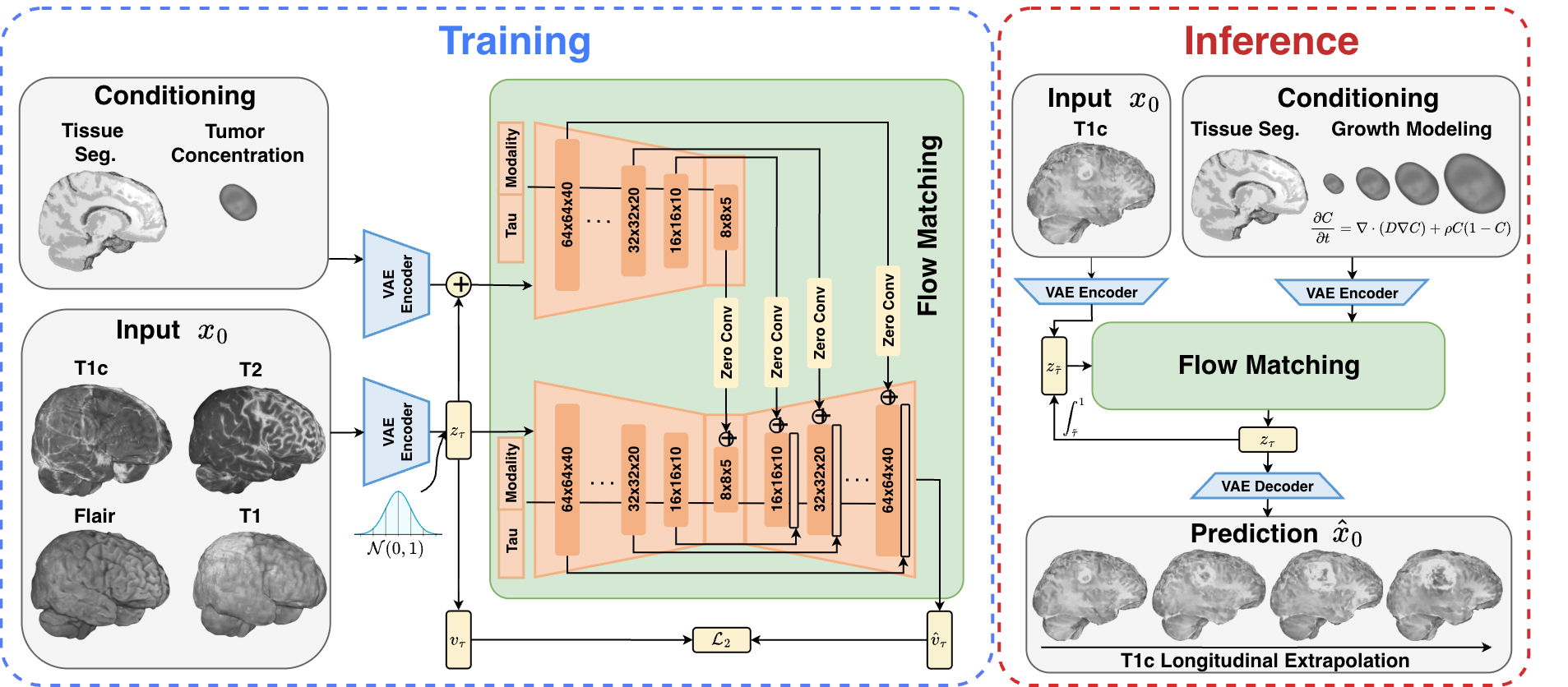}
\caption{\textbf{Architecture of TumorFlow.} Modality, tissue segmentations, and tumor concentrations are provided as conditioning inputs to the latent generative model. Orange boxes denote trainable modules, whereas blue boxes indicate frozen, pre-trained components. During inference, a full growth trajectory can be extrapolated from a biophysical simulation.}
\label{fig:main} 
\end{center}
\end{figure}

We present a two-stage 3D latent generative framework in Fig.~\ref{fig:main} for anatomically consistent brain tumor MRI synthesis and progression modeling. A pretrained 3D VAE encodes multimodal MR volumes, and a flow matching model in latent space conditions on modality, tissue segmentation, and biophysically-derived tumor-concentration fields. During inference, temporally propagated concentration maps enable biologically realistic tumor growth prediction.

\textit{\textbf{Autoencoding.}}
Let $\mathbf{x} \in \mathbb{R}^{256\times256\times160}$ denote a single-modality 3D MRI volume. We utilize a pretrained variational autoencoder (VAE) from the MAISI framework~\cite{guo2025maisi} which
maps $\mathbf{x}$ into latent space $\mathbf{z}$ given the encoder $\mathcal{E}_\phi$ approximating the posterior distribution \(q_\phi(z|x)\) with \(z \sim \mathcal{N}(\mu_z, \Sigma_z)\)
from which latent codes $\mathbf{z}\in\mathbb{R}^{4\times64\times64\times40}$ 
are sampled via the reparameterization trick. Afterwards, the image can be reconstructed again by estimating the likelihood \(p_\theta(x|z)\) using the decoder $\mathcal{D}_\theta$, where $\phi$ and $\theta$ are learnable parameters. 

\textit{\textbf{Latent Flow Matching.}}
For generation, we utilize an Optimal Transport Flow Matching (OT-FM) formulation. Let $\mathbf{z}_1$ be a latent from a real MRI via the VAE encoder and $\mathbf{z}_0 \sim \mathcal{N}(\mathbf{0},\mathbf{I})$ a Gaussian prior. We define an affine interpolation between source and target as
$
\mathbf{z}_\tau = (1-\tau)\mathbf{z}_0 + \tau \mathbf{z}_1
$
with $\tau \in [0,1]$
and ground-truth velocity
$
\frac{d\mathbf{z}_\tau}{d\tau} = \mathbf{z}_1 - \mathbf{z}_0.
$
\noindent A neural network $v_\psi(\mathbf{z}_\tau, \tau, \mathbf{c})$ with learnable parameters $\psi$ is conditioned on $\mathbf{c}$ to predict the velocity field such that training minimizes
\begin{equation}
\mathcal{L}_{\text{FM}}(\psi) 
= \mathbb{E}_{\mathbf{z}_0, \mathbf{z}_1, \tau, \mathbf{c}}
\left[
\left\|
v_\psi(\mathbf{z}_\tau, \tau, \mathbf{c}) - (\mathbf{z}_1 - \mathbf{z}_0)
\right\|_2^2
\right],
\quad \tau \sim \mathcal{U}[0,1].
\end{equation}

\noindent At inference, a new sample $\hat{\mathbf{z}}_1$ can be generated by drawing from source and integrating over the learned velocity field.
$
\hat{\mathbf{z}}_1 = \mathbf{z}_0 + \int_0^1 v_\psi(\mathbf{z}_\tau, \tau, \mathbf{c})\, d\tau.
$
The resulting latent $\hat{\mathbf{z}}_1$ is decoded by the VAE to reconstruct the final volume.

\textit{\textbf{Conditioning.}}
The conditioning input $\mathbf{c} = (m, \mathbf{c}_s, \mathbf{c}_{\mathrm{tc}})$ comprises MRI modality $m$, tissue segmentation $\mathbf{c}_s$, and tumor concentration map $\mathbf{c}_{\mathrm{tc}}$. 
The modality embedding $\mathbf{e}_m = \mathrm{Embed}(m)$ is concatenated with the timestep embedding. 
Both $\mathbf{c}_s$ (from atlas registration) 
and $\mathbf{c}_{\mathrm{tc}}$ are encoded by the pretrained encoder $\mathcal{E}$, yielding 
$\mathbf{z}_s = \mathcal{E}(\mathbf{c}_s)$ and 
$\mathbf{z}_{\mathrm{tc}} = \mathcal{E}(\mathbf{c}_{\mathrm{tc}})$. 
Their concatenation forms the spatial conditioning tensor 
$
\mathbf{c}_{\mathrm{spatial}} = \mathrm{concat}(\mathbf{z}_s, \mathbf{z}_{\mathrm{tc}}) \in \mathbb{R}^{8 \times 64 \times 64 \times 40},
$
which is injected into the U-Net backbone via a ControlNet-style branch~\cite{zhang2023adding}:
$
\mathbf{h}_\ell \leftarrow \mathbf{h}_\ell + \mathcal{F}_\ell(\mathbf{c}_{\mathrm{spatial}}),
$
where $\mathcal{F}_\ell$ are learnable projections at scale $\ell$.

\textit{\textbf{Biophysical Tumor Growth Modeling.}}
We model glioblastoma growth using the Fisher-Kolmogorov reaction-diffusion equation, which is widely adopted in the field \cite{ezhov2023learn,balcerak2025individualizing,lipkova2019personalized}. It models the temporal change of the tumor concentration $C$ based on a reaction and a diffusion term:

\begin{equation}
  \frac{\partial C}{\partial t}
  \;=\;
  \nabla \cdot (D\,\nabla C )
  \;+\;
  \rho\,C\, (1 - C),
  \label{eq:fisher_kolmogorov}
\end{equation}

\noindent where $D$ is a tissue-dependent diffusion coefficient, and $\rho$ the proliferation rate. 

\textit{\textbf{Longitudinal Image Generation.}}
Longitudinal generation must balance the temporal consistency of patient-specific anatomy with the need to reflect progressive tumor growth.
Let $\{\mathbf{c}_{\mathrm{tc}}^{t}\}_{t=0}^T$ 
denote time-indexed $t$ tumor concentration fields. The initial concentration $\mathbf{c}_{\mathrm{tc}}^{0}$ is estimated by a biophysical growth model~\cite{weidner2024spatial} to fit the pre-op tumor segmentation.
Each $\mathbf{c}_{\mathrm{tc}}^{t}$ is encoded into latent space and used for conditional synthesis. For temporal coherence and synthesis of time $t+1$ the previous state $z^{t}_1$ gets transported backward along the deterministic flow trajectory into a corrupted version $z^{t}_{\tilde{\tau}}$. Given this intermediate representation, we then integrate forward under the updated conditioning, enabling controlled temporal evolution while preserving anatomical consistency:
\begin{equation}
\hat{\mathbf{z}}^{t+1}_1 = \mathbf{z}^{t}_{\tilde{\tau}} + \int_{\tilde{\tau}}^1 v_\psi(\mathbf{z}^{t}_\tau, \tau, \mathbf{c}^{t}_{tc})\, d\tau.
\end{equation}

This scheme is recursively applied, yielding anatomically and temporally consistent longitudinal MRI synthesis. The corruption level $\tilde{\tau}$ is selected empirically for each method.

\section{Experiments}
\begin{table}[t]
\centering
\setlength{\tabcolsep}{8pt}
\caption{Quantitative evaluation of generated volumes. Dice compares thresholded tumor concentration fields with segmentations from a pretrained BraTS model. FID, KID, and MS-SSIM assess distributional alignment, perceptual similarity, and diversity against the test set. The reference metrics are empirical upper bounds derived from test-set calibration experiments, reflecting limits by VAE reconstruction fidelity, uncertainty in tumor-concentration derivations, and automatic tumor segmentation.}
\begin{tabular}{l lcccc}
\toprule
& \textbf{Model} & \textbf{Dice $\uparrow$} & \textbf{FID $\downarrow$} & \textbf{KID $\downarrow$} & \textbf{MS-SSIM $\uparrow$} \\
\midrule
\rowcolor{gray!20}
     & Reference & 0.779 & 0.330 & 0.001 & -- \\
\midrule
  \textit{(Ours)}   &     TumorFlow    & \textbf{0.739} & \textbf{2.125} & \textbf{0.020} & \textbf{0.675} \\
  \textit{Baseline} & Med-DDPM~\cite{dorjsembe2024conditional} & 0.344          & 106.246        & 1.618          & 0.588          \\
\midrule
\multirow{2}{*}{\textit{Ablation}}
  & OT-FM w. ViT                & \underline{0.572} & \underline{2.540} & \underline{0.026} & \underline{0.675} \\
  & DDPM w. U-Net               & 0.579          & 11.221         & 0.083          & 0.648          \\
\bottomrule
\end{tabular}
\label{table:generation_metrics}
\end{table}

\textit{\textbf{Implementation Details.}}
We use the publicly available BraTS~2021 dataset~\cite{baid2021rsna} and additional cohorts such as UCSF-PDGM~\cite{calabrese2022university}, TCGA-GBM~\cite{scarpace2016cancer}, TCGA-LGG~\cite{pedano2016cancer}, and Rembrandt~\cite{sayah2022enhancing}, excluding overlapping patients. This aggregation increases distributional diversity~\cite{usman2025beware} and mitigates memorization risk~\cite{usman2024brain}. The combined dataset comprises 3{,}602 subjects, split into 80\% training (2{,}881) and 20\% validation (721). All volumes are co-registered to a common anatomical template, resampled to $1\,\mathrm{mm}^3$ isotropic resolution, and skull-stripped. For longitudinal evaluation, we include the Lumiere dataset~\cite{suter2022lumiere}.
Models are trained on a single NVIDIA A100 (40\,GB) using AdamW (learning rate $10^{-4}$, zero weight decay) with cosine annealing and an EMA decay of $\gamma=0.999$ for stability.

\textit{\textbf{References, Baseline and Ablations}}  
Evaluation uses the held-out test set as a calibration baseline. Unless stated otherwise, Fréchet Inception Distance (FID) and Kernel Inception Distance (KID) are reported in units of $\times10^{-3}$. Comparing reconstructions from the pretrained VAE yields PSNR~32.09 and FID/KID~0.330/0.001, establishing the empirical fidelity ceiling. For tumor modeling, Dice overlap between reconstructed concentration fields and MR-based automatic segmentations~\cite{kofler2025brats} reaches 0.779, reflecting the uncertainty-imposed upper bound on segmentation accuracy.

We compare against the BraTS-pretrained Med-DDPM baseline~\cite{dorjsembe2024conditional} by thresholding the grown tumor concentration map to obtain a segmentation mask for conditioning. Specifically, edema is assigned for concentration values in $[0.2, 0.6)$ and enhancing tumor for values $\geq 0.6$.

We further compare against two ablations: 1. switching the corruption para-digm to a Denoising Diffusion Probabilistic Model (DDPM), and 2. replacing the 3D U-Net backbone with a Vision Transformer (ViT)~\cite{dosovitskiy2020image} using its base configuration and patch size of 2.


\begin{figure}[t]
        \centering
        \includegraphics[width=1\textwidth]{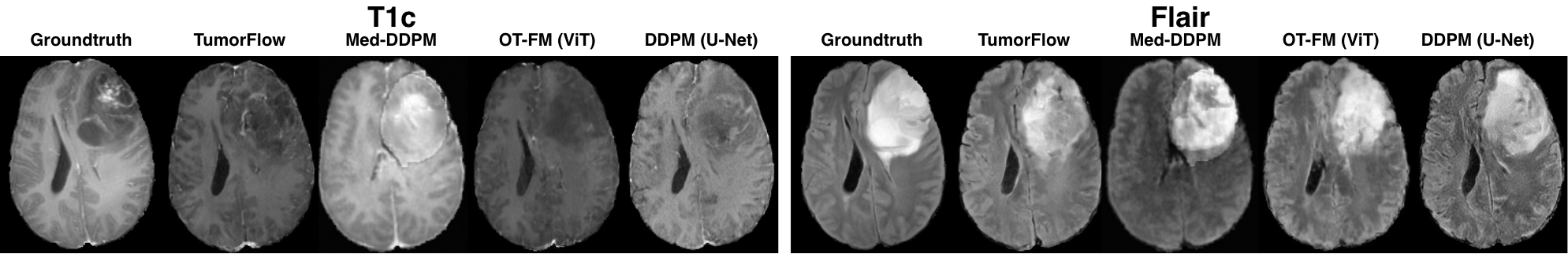}
        \captionof{figure}{\textbf{Qualitative Comparison}. Reconstructions from different methods.}
        \label{fig:static_qualitative} 
\end{figure}

\begin{figure}[t]
    \centering
    \includesvg[width=\textwidth]{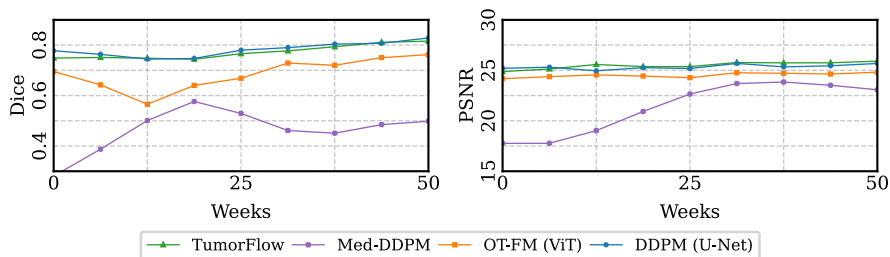}
    \captionof{figure}{
    \textbf{Left: Tumor Consistency.} Dice across time steps, assessing tumor growth consistency. \textbf{Right: Tissue Consistency.} PSNR across time steps measuring structural drift in healthy tissue. Stable values indicate coherent generation.}
    \label{fig:spatio_temporal_weeks}       
\end{figure}

\textit{\textbf{Generation of Static Brain Tumor MRI.}} We first assess the general \textit{generative performance} of the model by evaluating image fidelity and diversity on the static, non-longitudinal setup. Specifically, we compute FID, KID, and MS-SSIM between the generated volumes and the real test volumes to quantify distributional fidelity and diversity of synthesized samples. Second, we evaluate \textit{adherence} to the conditioning signal by comparing tumor volumes derived from the input concentration field and from the generated images. The reference volume is obtained by thresholding the concentration field at $0.2$, which is the calibration threshold for the biophysical growth model. Tumor volumes in the synthesized images are estimated using an independent pretrained segmentation model within the BraTS Orchestrator~\cite{kofler2025brats}, based on the top-performing method from the Adult Glioma Segmentation Challenge~\cite{baid2021rsna}. Adherence is quantified using the Dice Score~\cite{zou2004statistical} between the thresholded conditioning mask and the predicted segmentation. 

Quantitative evaluation of fidelity, diversity, and conditioning adherence de-monstrates that our model outperforms the Med-DDPM~\cite{dorjsembe2024conditional} baseline as well as the diffusion and ViT ablations in brain MRI synthesis (Table~\ref{table:generation_metrics}). Lower FID and KID scores indicate a closer match to real data, while higher MS-SSIM values confirm improved perceptual and structural fidelity. Overall, the Flow Matching formulation produces more realistic and anatomically consistent volumes than diffusion-based counterparts.
Additionally, our model demonstrates substantially improved \textit{tumor adherence} compared to the Med-DDPM~\cite{dorjsembe2024conditional} baseline, with generated tumor regions aligning more closely to the intended geometry, as shown qualitatively in Fig.~\ref{fig:static_qualitative}. In particular, leveraging tumor concentration maps leads to higher Dice similarity between the target signal and the synthesized tumor, reflecting tighter spatial agreement.

\begin{figure*}[t]
    \centering
    \begin{minipage}[t]{0.6\textwidth}
        \vspace{0pt}
        \centering
        \includegraphics[width=1.0\textwidth]{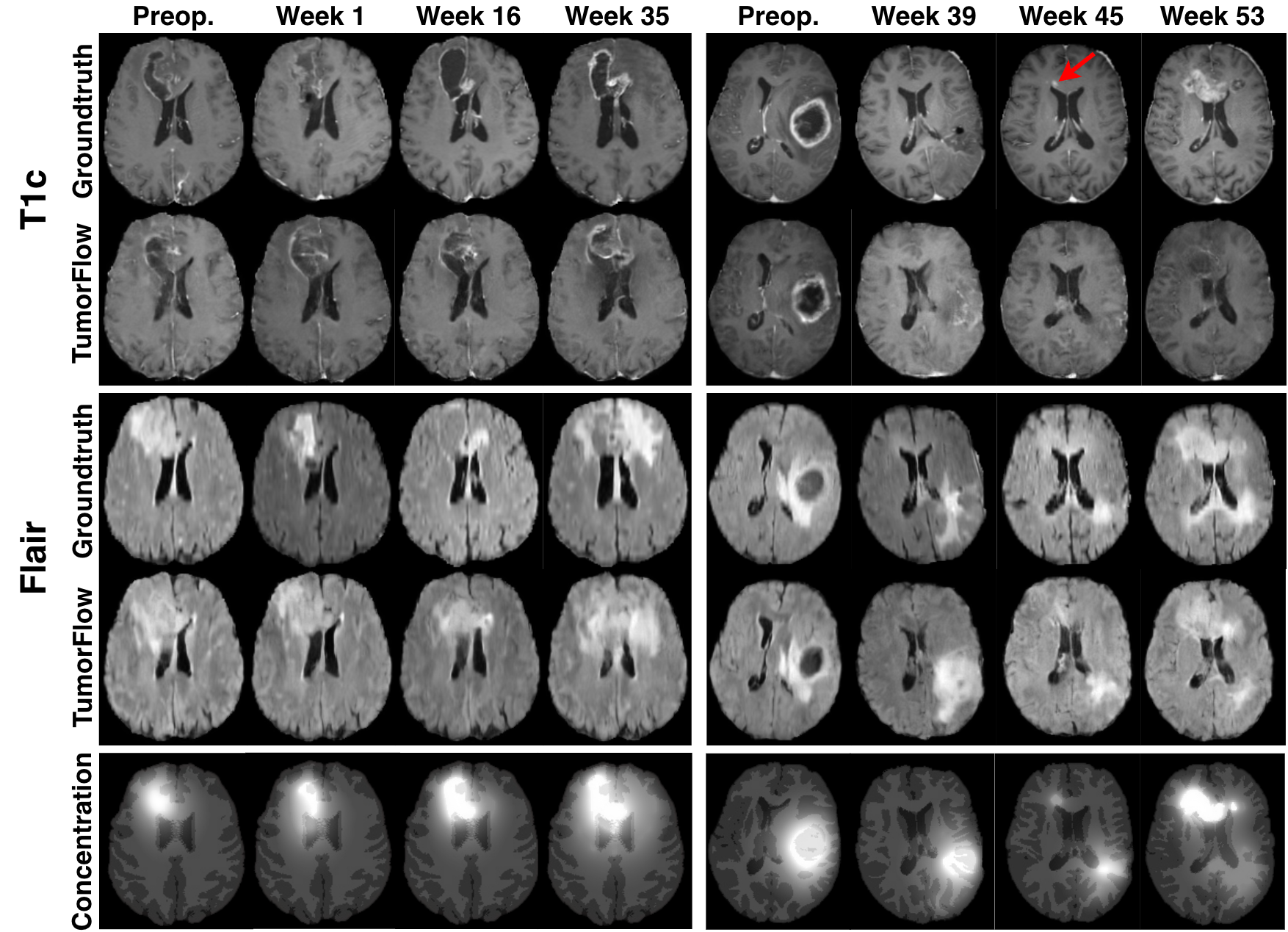}
        \label{fig:lumiere_a}
    \end{minipage}
    \hfill
    \begin{minipage}[t]{0.38\textwidth}
        \vspace{0pt}
        \centering
        \includesvg[width=1.0\textwidth]{figures/spatio_temporal_noise.svg}
        \label{fig:spatio_temporal_corruption}
    \end{minipage}
    \caption{\textbf{Left: Real patients.} Comparing TumorFlow and groundtruth tumor development in two patients from Lumiere \cite{suter2022lumiere}. The red arrow points to a new tumor lesion in the corpus callosum. \textbf{Right: Effect of corruption initialization on longitudinal generation.} Top: Dice between generated tumors and conditioning. Bottom: PSNR between consecutive non-tumorous regions.}
    \label{fig:lumiere}
\end{figure*}


\textit{\textbf{Longitudinal Extrapolation.}} The TumorGrowthToolkit~\cite{balcerak2025individualizing} is configured with proliferation rate $\rho = 0.03 \, \mathrm{d}^{-1}$ and diffusion coefficient $D = 0.28 \, \mathrm{mm}^2 \mathrm{d}^{-1}$  to generate physics-informed spatiotemporal tumor concentration fields ${c^{t}(\mathbf{x})}_{t=0}^{T}$, from which corresponding longitudinal MRI sequences are synthesized. \textit{Temporal consistency} is enforced via predecessor-manipulation, where each time point is conditioned on the evolved concentration field and initialized at $\tilde{\tau}=0.15$.

In Fig.~\ref{fig:synthetic_progression}, we show a representative longitudinal prediction over 50 weeks.
The biophysical simulation produces a smooth tumor cell concentration trajectory that informs the generation of follow-up scans. 
Despite substantial changes in the extent of the tumor, the patient anatomy outside of the tumor remains stable over time. The growing tumor, in particular the necrotic and contrast-enhancing areas, exhibits realistic morphology and imaging characteristics. The downstream segmentations provide an additional consistency check, producing masks that evolve plausibly and align well with the predicted growth pattern.

In the left panel of Fig.~\ref{fig:spatio_temporal_weeks}, \textit{temporal consistency} of the simulated tumor progression is evaluated using the Dice score, where our model achieves the highest temporal stability, indicating that the evolving, synthetic tumors are highly realistic. 
On the right panel of that figure, temporal tissue consistency is quantified by computing pixel-wise fidelity in non-tumorous regions between consecutive time points. Structural drift in healthy tissue is measured by PSNR: excessively high values would indicate an overly static anatomy that impedes realistic tumor growth, while low values reflect excessive structural drift and diminished subject coherence over time. Our model achieves a balanced trade-off, maintaining stable Dice and PSNR curves that indicate temporally coherent synthesis. Over time, consistent metric values suggest controlled tumor progression and preserved anatomy, whereas deviations reveal either insufficient tumor growth dynamics or instability in structural tissue fidelity.

As treatment-naive longitudinal public data do not exist, we evaluate our method on the postoperative Lumiere Dataset (see Fig.~\ref{fig:lumiere} left). The images contain a resection cavity and post-surgery brain deformation, for which our model is not trained. We estimate tumor concentration by independently fitting a biophysical model at each time step. Generation is initialized from the pre-op image.
Despite the resection cavity, we observe temporally consistent and anatomically plausible longitudinal images. In the right patient, we note a satellite lesion in the corpus callosum at week 45 (red arrow). We observe that this satellite lesion closely follows the predicted concentration growth pattern, while the surrounding anatomy remains visually consistent.

\textit{\textbf{Ablation.}}  
In the stepwise corruption process of the optimal transport formulation, fine-grained features degrade earlier than coarse structures. To assess the impact of this, we evaluate our model and its variants across different architectures by varying the corruption level $\tilde{\tau}$ used to initialize the predecessor manipulation on the right panel of Fig.~\ref{fig:lumiere}.
Low \( \tilde{\tau} \) values yield volumes nearly identical to the previous state, resulting in high PSNR but minimal tumor progression. Increasing \( \tilde{\tau} \) induces stronger updates and more realistic growth dynamics. Our TumorFlow model exhibits a smooth, monotonic Dice response across corruption states, indicating stable and precise controllability of tumor evolution. In contrast, DDPM-based models exhibit irregular, less predictable trends, reflecting a weaker balance between anatomical stability and tumor fidelity. Med-DDPM~\cite{dorjsembe2024conditional} further deviates from the intended progression.

\section{Conclusion}
We introduced a physics-guided generative framework for longitudinal glioblastoma MRI synthesis that conditions high-fidelity 3D image generation on spatially continuous tumor concentration fields and evolves these fields over time with a Fisher-Kolmogorov reaction-diffusion growth model. 
By conditioning on spatially continuous tumor concentration fields and evolving them with a biophysical growth prior, we can generate follow-up images that reflect biologically plausible tumor growth trajectories while keeping the individual anatomy and image appearance stable over time. Importantly, we achieve this while training only on cross-sectional, preoperative data, effectively alleviating the persistent shortage of high-quality longitudinal datasets in medical imaging.


Future work will focus on more realistic modeling of tumor growth and treatment response, where our work lays an important foundation for \textit{in silico} disease and treatment modeling.


%
%
%
\newpage
\bibliographystyle{splncs04}
\bibliography{references}

\end{document}